\documentclass[10pt,twocolumn,letterpaper]{article}

\usepackage{iccv}
\usepackage{times}
\usepackage{epsfig}
\usepackage{graphicx}
\usepackage{amsmath}
\usepackage{amssymb}
\usepackage{algorithm}
\usepackage{algpseudocode}

\usepackage[breaklinks=true,bookmarks=false]{hyperref}

\iccvfinalcopy 

\def\iccvPaperID{8787} 

\ificcvfinal\fi

\begin{document}

\title{Demystifying the Effects of Non-Independence in Federated Learning}

\author{Stefan Arnold\\
Friedrich-Alexander-University\\
Erlangen-Nuremberg\\
Institute of Information Systems\\
90403 Nuremberg, Germany\\
{\tt\small stefan.st.arnold@fau.de}
\and
Dilara Yesilbas\\
Friedrich-Alexander-University\\
Erlangen-Nuremberg\\
Institute of Information Systems\\
90403 Nuremberg, Germany\\
{\tt\small dilara.yesilbas@fau.de}
}

\maketitle
\ificcvfinal\thispagestyle{empty}\fi

\begin{abstract}
Federated Learning (FL) enables statistical models to be built on user-generated data without compromising data security and user privacy. For this reason, FL is well suited for on-device learning from mobile devices where data is abundant and highly privatized. Constrained by the temporal availability of mobile devices, only a subset of devices is accessible to participate in the iterative protocol consisting of training and aggregation. In this study, we take a step toward better understanding the effect of non-independent data distributions arising from block-cyclic sampling. By conducting extensive experiments on visual classification, we measure the effects of block-cyclic sampling (both standalone and in combination with non-balanced block distributions). Specifically, we measure the alterations induced by block-cyclic sampling from the perspective of accuracy, fairness, and convergence rate. Experimental results indicate robustness to cycling over a two-block structure, e.g., due to time zones. In contrast, drawing data samples dependently from a multi-block structure significantly degrades the performance and rate of convergence by up to $26$\%. Moreover, we find that this performance degeneration is further aggravated by unbalanced block distributions to a point that can no longer be adequately compensated by higher communication and more frequent synchronization.
\end{abstract}

\section{Introduction}

\textit{Federated Learning} (FL) was introduced by McMahan \etal~\cite{mcmahan2017communication} with an initial emphasis on applications for mobile devices, \eg, keyboard prediction \cite{hard2018federated,ramaswamy2019federated,yang2018applied}. Following the recent interest in using FL for other applications  \cite{li2020review}, it has emerged as the leading paradigm to address the omnipresent conflict between utility and privacy of user-generated data \cite{kairouz2019advances}.

FL poses challenges that differ fundamentally from distributed optimization \cite{hu2018stochastic}. In particular, FL is exposed to statistical heterogeneity. When referring to statistical heterogeneity, it means that user-generated data residing on mobile devices usually follows a non-representative distribution of the total population. This statistical heterogeneity is violating the fundamental assumptions of independent and identical distributions (i.i.d.). Data non-compliance with i.i.d. (non-i.i.d.) has been observed to induce a divergence to the model weights which may significantly degrade the performance and convergence of FL \cite{zhao2018federated}. The problem of non-i.i.d. data is further complicated by the fact that the data in practical FL environments are massively and unbalanced distributed across devices.

In this study, we respond to the recent call of Kairouz \etal \cite{kairouz2019advances} to extend the current understanding of the non-i.i.d. and unbalanced properties of FL. Most studies in terms of statistical heterogeneity have been devoted to challenges of non-identical data distributions \cite{hsu2019measuring,zhao2018federated} or non-balanced data distributions \cite{duan2020self,wang2020towards,yang2020federated}. Although non-independence is a common phenomenon in FL, mainly due to temporal unavailability of mobile devices \cite{eichner2019semi}, almost no research is concerned with non-independent data distributions. For this reason, we focus on quantifying and understanding the implications brought by block-cyclic sampling with and without additional block imbalance. We particularly emphasize on accuracy parity \cite{zafar2017fairness} among clients \cite{mohri2019agnostic,li2019fair}.

The remainder of this study is organized as follows: In Section \ref{section:2}, we elaborate on the theoretical background of statistical heterogeneity and briefly review related work. In Section \ref{section:3}, we explain the experimental setup that has been used to simulate and analyze block-cyclic sampling. This includes data preparation and the manipulation of the vanilla FL selection protocol. In Section \ref{section:4}, we present the results obtained given both a fixed and fine-tuned optimization budget. Our empirical results help to illustrate and validate the performance of FL in applications consisting of heterogeneous blocks. In Section \ref{section:5}, we conclude our research by advocating carefully tuned hyperparameters and call for the development of novel algorithmic subroutines to address the problem of data dependence.

\section{Background} \label{section:2}

Coordinated by a central parameter server, FL builds a statistical model using an iterative multi-step protocol. Each iteration constitutes a round of communication between the central server and the devices at the edge, also known as \textit{clients}. At the beginning of the protocol, a statistical model with a fixed model structure is compiled using a generic initialization $\theta$. In each round, a random subset of clients is selected for improving the global model. To improve the global model, the server broadcasts a copy of the global model to a randomly selected portion of available clients. Every client then trains its copy by performing multiple local updates based on their local data. To perform local updates, stochastic gradient descent (SGD) is typically applied. Upon completing a fixed number of local updates, clients report the changes made to the copy of the global model back to the central server. This report takes the form of a \textit{d}-dimensional vector of model weights. After aggregating the local updates from the selected subset of clients, the central server calculates a weighted average of the received model weights to obtain an optimized global model. The vanilla averaging procedure in FL is termed \texttt{FedAvg}. Implementation details of \texttt{FedAvg} are shown in Algorithm \ref{alg:Algorithm 1}. The aggregation concludes one round of communication. By dispatching the updated global model to another random subset of clients, the process repeats itself until a fixed communication budget is depleted or convergence is achieved.

\begin{algorithm}
\caption{Federated Averaging. The $K$ clients are indexed by $k$; $B$ is the local mini-batch size, $E$ is the number of local epochs, and $\eta$ is the learning rate.}
\label{alg:Algorithm 1}

\begin{algorithmic}[1]
\State $\text{global model } w_{0}$

\Procedure{ServerUpdate}{}
\For{$\text{each round } t \in [T]$}
\State $m \gets \max(C \times K,1)$
\State $S_{t} \gets \text{random subset of } m \text{ with } p_{k}=\frac{1}{K} $
\For{$\text{each client } k \in S_{t}$}
\State $w^k_t+1 \gets \text{ ClientUpdate}(k,w_{t})$
\EndFor
\State $ w_{t} \gets \sum_{k=1}^{K} \frac{n_{k}} {n \times w^{k}_{t+1}}$
\EndFor
\EndProcedure

\Function{ClientUpdate}{k,w}
\State $B \gets \text{split local data into batches of size } b$
\For{$\text{each local epoch } i=0,\ldots ,E-1$}
\For{$\text{batch } b \in B $}
\State $w \gets w + \eta \times \nabla l(w;b)$
\EndFor
\EndFor
\State \textbf{return} $w$
\EndFunction

\end{algorithmic}
\end{algorithm}

While i.i.d. data is formally defined, non-i.i.d. data properties are manifested by different forms. To ensure a consistent understanding of non-i.i.d. data regimes, we provide a taxonomy of non-i.i.d. data in reference to Hsieh \etal~\cite{hsieh2020non} and Kairouz \etal~\cite{kairouz2019advances}. The taxonomy adopts notions of dataset shift \cite{moreno2012unifying}. However, instead of focusing on differences between training and test data distribution, their taxonomy considers differences in the data distribution between clients.

Building a statistical model in a supervised manner using FL involves two stages of sampling: (i) selecting a client $i \sim Q$, where $Q$ denotes the distribution over available clients, and (ii) drawing a sample-label pair from the local data distribution, i.e., $(x,y) \sim P_i(x,y)$, where $P_i$ denotes the data distribution of the selected client $i$.
 
In FL, dependency mostly arises from the fact that each client corresponds to a user with specific interests, geographic location, and time zone. Therefore, the principle of independence is violated whenever the distribution $Q$ changes over the course of training. Due to the design of FL on mobile devices, only devices that are idle, charging, and connected to an unmetered wireless connection are available for selection. Noted that client availability follows temporal and/or behavioral patterns, therefore this condition is typically satisfied during nighttime hours. As a consequence, the distribution $Q$ changes cyclically.

In contrast to the violation of independence, common ways in which the principle of identicalness is violated are manifold. In FL, we distinguish between covariate shift, prior probability shift and concept shift. Covariate shift, also referred to as feature distribution skew, indicates the scenario in which the marginal distributions $P_i(x)$ varies across clients even if $P(y \vert x)$ is shared. This is the case, for example, with handwriting recognition, where clients write the same words with different accentuations, such as stroke widths. Prior probability shift, also referred to as label distribution skew, indicates the scenario in which the marginal distributions $P_i(y)$ varies across clients even if $P(x \vert y)$ is shared. For example, clients located in different geographic regions may contain images of regional animals. Due to differences in the demographics of clients, skewed distribution of labels is most common for user-generated data. Thus, majority of research in FL is devoted to dealing with skewed labels. Concept shift may also be a reason for data to deviate from being identically distributed. It occurs in two distinct settings. First, the conditional distributions $P_i(x \vert y)$ may vary across clients even if $P(y)$ is shared. In this case, the same label can have very dissimilar features for different clients. For example, images of clothing may vary widely across geographic regions, due to culture, weather conditions, standards of living or whether the images were captured during the day or at night. Second, the conditional distributions $P_i(y \vert x)$ may vary across clients even if $P(x)$ is shared. In this case, the same features can be interpreted as different labels due to personal preferences. For example, next word prediction may vary across regions and depending on the client’s personal language. Similar, labels that reflect the sentiment of sentences have personal and regional variation. Apart from skewed distributions, clients may generate and hold vastly different amounts of data. This case resembles a form of data imbalance.

McMahan \etal~\cite{mcmahan2017communication} demonstrated by means of experiments that FL is able to converge under non-identical data distributions. Lacking a theoretical proof of convergence, Li \etal\cite{li2018federated} complemented theoretical convergence guarantees. Li \etal\cite{li2019convergence} and Khaled \etal~\cite{khaled2020tighter} further tightened the theoretical guarantees for the convergence of distributed optimization on statistically heterogeneous data.

Assuming that data are drawn independently from differing distributions, the impact of non-identically distributed data has been highlighted by several studies \cite{hsieh2020non,hsu2019measuring,zhao2018federated}. Zhao \etal~\cite{zhao2018federated} attribute the detrimental effect to weight divergence, which refers to an observed drift in local models due to on-device training based on non-representative data. As the number of local training iterations increases, the drifts become more pronounced yielding a higher generalization gap. Recently, this line of work was extended upon by Chen \etal~\cite{chen2020fedmax}. The authors attribute the performance degeneration to a divergence in the activation in addition to the divergence in weights. The activation in neural networks is related to the last dense layer to yield logits, \ie, unnormalized probabilities.
 
To address the issue of non-identicalness, several algorithmic extensions have been proposed \cite{duan2020fedgroup,karimireddy2020scaffold,li2018federated,shoham2019overcoming}. Zhao \etal~\cite{zhao2018federated} propose sharing a small portion of the data between the participating clients. While this approach helped mitigating the detrimental effect of non-identicalness, it is contradictory with the fundamental privacy-sensitive learning objective of FL. Li \etal~\cite{li2018federated} introduce a proximal term to constrain the distance between local models and the global model. Showing that non-identical data leads to degradation of accuracy and a slowdown in the rate of convergence, Sattler \etal~\cite{sattler2019robust} advocate sparsification to permit more frequent global aggregation instead of local training iterations. Karimireddy \etal~\cite{karimireddy2020scaffold} suggest utilizing a form of stochastic variance reduction to control the local drifts caused by non-identical data distribution. Recently, Duan \etal~\cite{duan2020fedgroup} propose clustering the clients into multiple groups based on their cosine similarity of their local updates. By grouping clients into shared groups, a singular model is cascaded through intra-group and inter-group aggregation. Instead of increasing the resilience of a singular model to non-identical data, another line of research considers exploiting the statistical heterogeneity in non-identical data to construct pluralistic models through personalization. To attain personalization across groups of clients, several techniques from domain adaptation \cite{ben2010} can be employed, such as local fine-tuning \cite{arivazhagan2019federated,wang2019federated}, mixture of experts \cite{hanzely2020federated,zec2020federated}, meta-learning \cite{chen2018federated,fallah2020personalized}, or multi-task learning \cite{corinzia2019variational,smith2017federated}. We refer to Kulkarni \etal~\cite{kulkarni2020survey} for a comprehensive survey on personalization techniques in FL.

Compared to the plethora of research on non-identical data distributions, research on addressing the violations of independence stemming from changes in client availability is almost non-existent. For the purpose of language models, Eichner \etal~\cite{eichner2019semi} studied the performance and convergence rate of FL when multiple blocks of clients with differing characteristics are sampled according to a regular cyclic pattern (\ie, based on time zones). The authors show that block-cyclic sampling can significantly deteriorate the rate of convergence in FL. To mitigate the slowdown induced by the block-cyclic sampling, the authors suggest using pluralistic models. These models are built based on a semi-cyclic method that performs a one-time model averaging over the blocks of clients. In contrast to Eichner \etal~\cite{eichner2019semi}, our paper takes a different angle by focusing on visual classification instead of natural language processing. We also investigate the effect of cycling over multi-block structures while considering both balanced and unbalanced block distributions.

Induced by violations of balancedness, several algorithmic subroutines to FL have been proposed \cite{duan2020self,sarkar2020fed,wang2020towards,yang2020federated}. To mitigate the effect of skewed data distributions, Duan \etal~\cite{duan2020self} build a self-balancing framework that alleviates imbalance by applying a $z$-score-based data augmentation along with a multi-client rescheduling scheme. Yang \etal\cite{yang2020federated} developed an estimation scheme of the class distribution that is used to adjust the scheduling according to the class distribution. Wang \etal~\cite{wang2020towards} and Sarkar \etal~\cite{sarkar2020fed} modify the objective function by adapting focal loss \cite{lin2017focal}.

Reviewing related studies on  statistical heterogeneity yield that non-i.i.d. data regimes are typically examined in isolation with an apparent prevalence on non-identical data distributions. Highlighted by the recent findings of Eichner \etal~\cite{eichner2019semi}, we complement on the scarce body of research on non-independence. More specifically, we make the following contributions: (i) we measure the implications of non-independent data arising from block-cyclic sampling, (ii) we a provide generalization to multi-block structures, (iii) we measure the moderation effect of block imbalance, and (iv) we analyze the sensitivity to the choice of the learning rate. By conducting extensive experiments on the interactions of non-independence and non-balancedness, we advocate for a paradigm shift towards other regimes of non-i.i.d. data and their combinations.

\section{Methodology} \label{section:3}

Unlike the definitional perspective of the previous section, we now adopt an applied perspective to present the experimental setup. At the level of block structures, the experiments consist of three dimensions: (i) non-independence, (ii) non-identicalness, and (iii) non-balancedness. With rigorous experimental methodologies, we explore the effect of these dimensions on the vanilla \texttt{FedAvg} algorithm from the perspectives of accuracy, fairness, and convergence.

The MNIST dataset provides the basis for all pathological experiments. We use MNIST for reference as it is commonly used in FL literature. In total, the MNIST dataset contains $60,000$ labeled images from $10$ classes whereby additional $10,000$ images are reserved has holdout for validation. For the i.i.d. setting, we subsample a population of homogeneous clients with an almost equal number of samples per class. This represents an unrealistic setting since practical FL would typically involve clients with data generated from heterogeneous interests and behaviors. To synthetize this homogeneous population, we randomly assign a uniform distribution over all classes to each client without replacement. Therefore, each client holds a similar amount of data samples from all classes. 

In contrast to the i.i.d. setting, we use the term non-i.i.d. data to refer to samples that have been purposely redistributed to a client. To synthesize a population with non-i.i.d. data, we follow the approach of sorting and partitioning data into shards as put forward by McMahan \etal~\cite{mcmahan2017communication}. This approach is widely used to analyze and benchmark FL \cite{hsieh2020non,nilsson2018performance,sattler2019robust,zhao2018federated}. The step-by-step process is to sort the data by labels, divide the data label-wise into equally sized shards, and then randomly assign a fixed number $s \geq 1$ of shards to each client. Although this represents a pathological extreme case of non-identicalness, practical FL is likely to involve more complex distributions than partitions \cite{hsu2019measuring}.

\begin{figure}[t]
\centering
\includegraphics[scale=0.5]{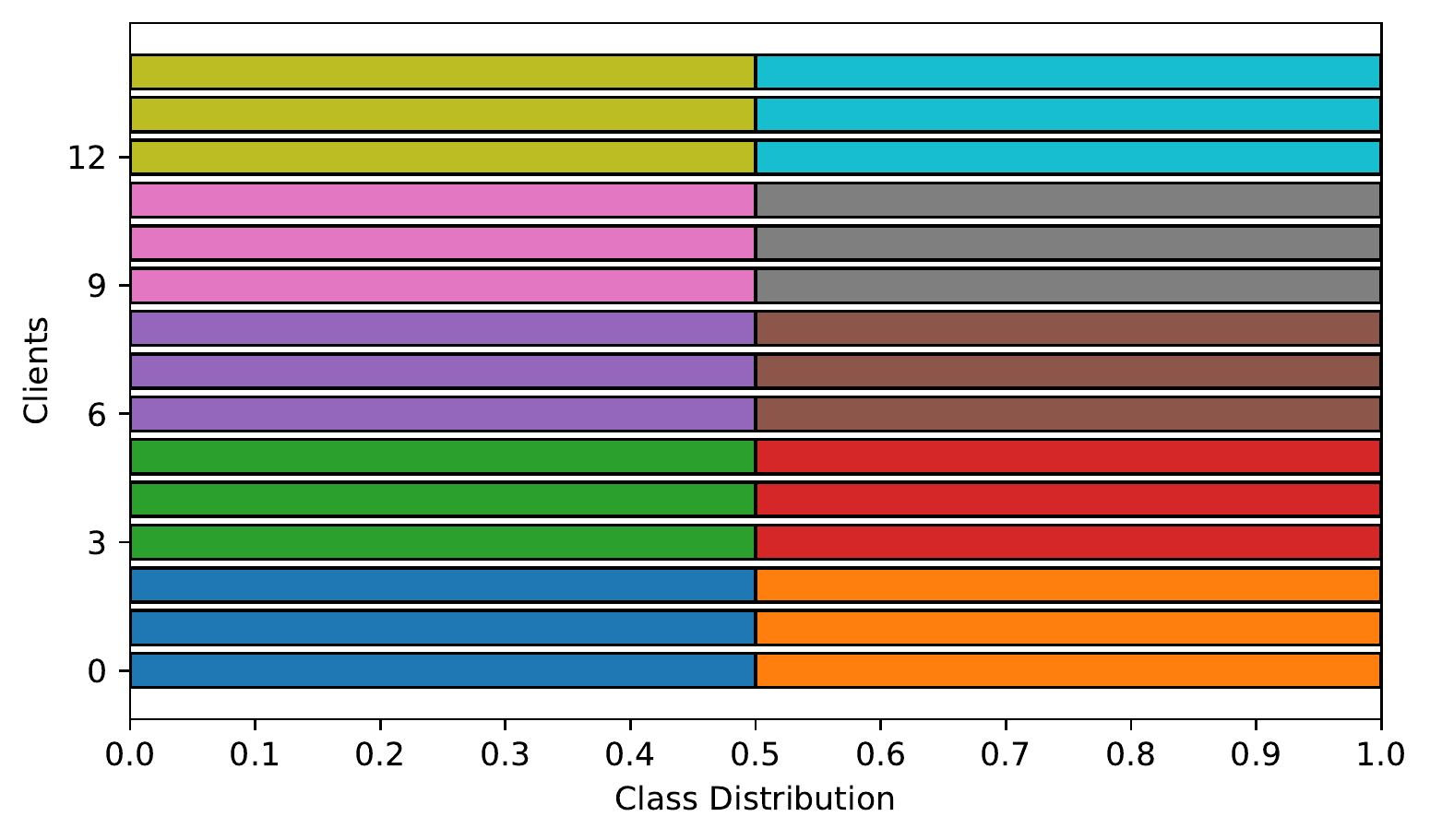}
\caption{Exemplary representation of a five-block structure with $2$ classes per block across $15$ clients. Classes are colored by label. Note that classes are not shared between the blocks.}
\label{fig:Figure 1}
\end{figure}

We modify this sort-and-partition manner to account for $n$-block structures. Instead of assigning a number of shards from all labels to each client at random, we group the sorted labels according to a configured number of equally sized blocks $G=\lbrace 2,5 \rbrace$. FL assumes $G=1$ by design, which means that the samples are drawn independently from a single homogeneous group consisting of all clients. Considering that samples from a block are processed each $G^{\text{th}}$ round, higher $G$ naturally yields to more dependence. Thus, any $G>1$ represents non-independence. Each client is then assigned to a block and subsamples shards associated with this block. We fixed the number of shards to $s=2$. With this modification we generate blocks consisting of non-overlapping classes. Figure \ref{fig:Figure 1} shows an example for an instantiation of this process with fifteen clients partitioned into a five-block structure. Note that classes (coded by color) are not shared between the blocks.

To account for a block-level non-balancedness, instead of assigning an equal number of samples per block, we distribute shards with replacement in a way that the number of samples across blocks follows a power law distribution. Note that this has a similar effect to distributing the number of clients per block following the power law. For the power law distribution, we experiment with a discrete range of scaling factors $\alpha \in \lbrace 1.0,1.5,2.0,5.0 \rbrace$. For example, a two-block structure corresponds to minority-majority ratio of approximately $\lbrace 1:1, 1:2, 1:3, 1:11 \rbrace$.
 
We separate a part of each client's data as a holdout set. To generate the local holdout sets, we subsample examples from the MNIST validation set using the same distribution that has been used to construct the local training data. Thus, the holdout set inherits the local distribution and therefore each client’s test data is different. In each round, we measure and report the accuracy of the global model on the union of the holdout set and the local accuracy of each client on its own local holdout data. In this way, we account for the recent debate on fairness in FL \cite{huang2020fairness,li2019fair,mohri2019agnostic}. To assess the fairness change induced by cyclic sampling, we adopt the definition of fairness formulated by Li \etal~\cite{li2019fair}. A consensus model $w$ is considered fair when its generalization performance $\lbrace A_1,…,A_m \rbrace$ is \textit{uniform} across $m$ clients.

In line with Eichner \etal~\cite{eichner2019semi}, we model and operationalize the block-structure by modifying the random selection of a subset of clients into block-cyclic selection. During the course of training, we cycle over blocks in a fixed and repeating order and select a configured fraction of clients belonging to the current block at random. For example, given two cyclically ordered blocks [A,B], we select a subset of clients $S_t=K \times C \leq \lceil G \rceil$ from block $A$ ($A \longrightarrow B \rightarrow A \longrightarrow \dotsb$ in consecutive rounds), where $K$ denotes the number of clients, $C \in [0,1]$ denotes the reporting fraction, and $\lceil G \rceil$ denotes the block size. Using $\lceil G \rceil$ limits the number of clients selected for training to the the number of clients per block

For the \texttt{FedAvg} algorithm, we use the same notations as McMahan \etal~\cite{mcmahan2017communication}. We simulate all experiments with a fixed setup regarding the batch size $B=64$, local epoch $E=3$, and number of clients $K=100$ indexed by $k$. We consider full and partial participation by allowing a variable fraction $C \in \lbrace 0.05,0.10,0.20,1.00 \rbrace$ (corresponding to $\lbrace 5,10,20,100 \rbrace$ clients participating in every communication round, respectively). To be consistent with \texttt{FedAvg}, the fraction is based on the total number of clients. In case the reporting fraction $K \times C$ exceeds the number of clients artificially assigned to a block, all clients from this block are chosen to participate. Thus, full participation translates into reporting fraction of $
\frac{K}{G}$ for all block-cyclic settings. \texttt{FedAvg} is run for a fixed optimization budget of $T=100$ communication rounds. No early stopping \cite{prechelt1998early} is applied. To obtain a statistically sufficient sample size that diminishes random variances, we substantiated our experiments by $n=3$ independent replications.

For our experiments, we exploit a simple convolutional neural network (CNN) architecture. The CNN model propagates data through two convolutional layers with filter size 3×3 and two dense layers. The first convolutional layer has $32$ channels, followed by maximum pooling with filter size $2 \times 2$ and dropout with probability $p=0.5$. The second convolutional layer has $64$ channels. Dropout is applied with $p=0.25$. The dense layers contain $128$ units. Except for the output layer, which uses a softmax to yield a probability distribution, all layers use rectified linear unit activation. In total the CNN model consists of $1,199,882$ trainable parameters. While this model is not state-of-the-art, it is sufficient to show relative performance for the purposes of our investigation. Note that the model is the same for each client. The model is initialized and transmitted to all clients. To optimize the model, we apply SGD with a fixed learning rate of $\eta =0.01$ and momentum  of $\beta =0.50$ \cite{sutskever2013importance}. We take the search for an optimal learning rate in the sensitivity analysis. For simplicity, we did not apply weight decay or any learning rate decay schedule.

\section{Experiments} \label{section:4}

For comparison, we consider a performance upper bound by creating a hypothetical i.i.d. case where images are distributed uniformly across $K=100$ clients, and training is performed each round $t$ by a subset of $\vert S_t \vert =20$ clients drawn uniformly at random. With this setup, we measure a test accuracy of $0.9901$. We also consider a more practical upper bound for the performance by sorting and distributing the images non-identically across clients. With the training remaining unchanged, we achieve a practical upper bound for the test accuracy of $0.9878$. This shows that FL  is capable to deal with non-identical data.

Given the above data preparation, we proceed to benchmark the \texttt{FedAvg} algorithm using block-cyclic sampling. We present our results on a discrete range of blocks ordered from balanced to unbalanced.

\subsection{Performance Analysis}

\begin{figure}[t]
\centering
\includegraphics[scale=0.5]{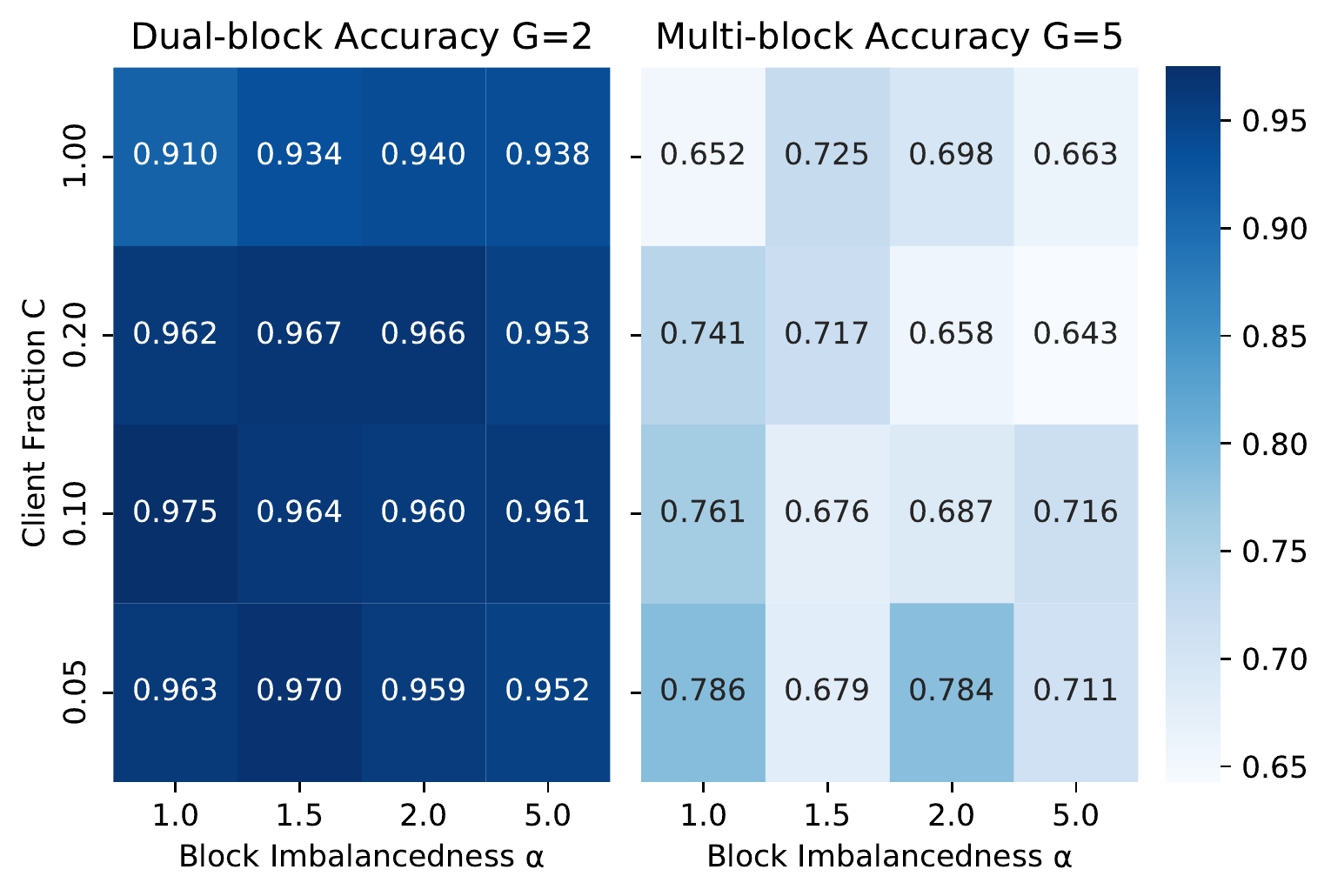}
\caption{\texttt{FedAvg} test accuracy for different degrees of non-independence and non-balancedness. Each cell is optimized with a communication budget of $T=100$ rounds synchronized every $E=3$ epochs. Each cell is averaged over $3$ seeded runs on different populations using the same $G$, $C$, and $\alpha$. }
\label{fig:Figure 2}
\end{figure}

We commence by presenting the performance impact of non-independence on non-identical data \footnote{Note that non-independence has no effect if data are identically distributed across all clients. This is because each client, regardless of the block, resembles a representative of the entire population.} given a fixed optimization budget. In Figure \ref{fig:Figure 2}, we depict the classification accuracy of the consensus model as a function of the power law scaling (larger implies more unbalanced group distributions) and client fraction, each grouped by number of blocks (larger implies more dependence). Significant changes in accuracy occur between blocks. Increasing the dependency by sampling from a multi-block structure $G \gg 2$ significantly degrades the performance by up to $0.253$ percentage points, on average. For $G = 2$, \texttt{FedAvg} yields a test accuracy of $0.9546$ on average, which is slightly below the practical upper bound. For $G = 5$, by contrast, \texttt{FedAvg} achieves a test accuracy of $0.7016$ on average for a range from $0.6426$ to $0.7865$. We conclude that the volatility of \texttt{FedAvg} increases disproportionately with the number of blocks $G$ for multi-block structures.

To investigate the moderating effect of the reporting fraction and ratio of imbalance within blocks, we calculate the per-block row-wise means (representing the effect of participation rates) and column-wise means (representing the effect of imbalance). While a wide range of fractions yield statistically indistinguishable test accuracies, we find that partial participation almost always leads to results than full participation. Recall that the client fraction in block-cyclic sampling is bound by $\frac{K}{G}$. Given this upper bound, the results of full participation in the five-block structure are not further considered. Findings regarding the moderating effect of imbalance in addition to dependence are inconclusive. While there is no significant difference between the test accuracy scores of balanced and unbalanced data for the two-block structure, we observe a more pronounced drop in test accuracy for the multi-block structure. For $G=5$, the accuracy degrades by up to $0.098$ percentage points. Despite the marginal effect in the case of the two-block compared to the five-block structure, we conclude that the implications of block-cyclic sampling are aggravated by unbalanced block distributions. This is concerning because unbalanced blocks may represent a realistic scenario when learning cyclically from differently engaged or sized geographic regions. For example, as the largest and most populous continent, rich sources of data are found in Asia. Due to cultural differences, however, the data found on mobile devices may differ greatly from those in America or Europe.

\begin{figure}[t]
\centering
\includegraphics[scale=0.7]{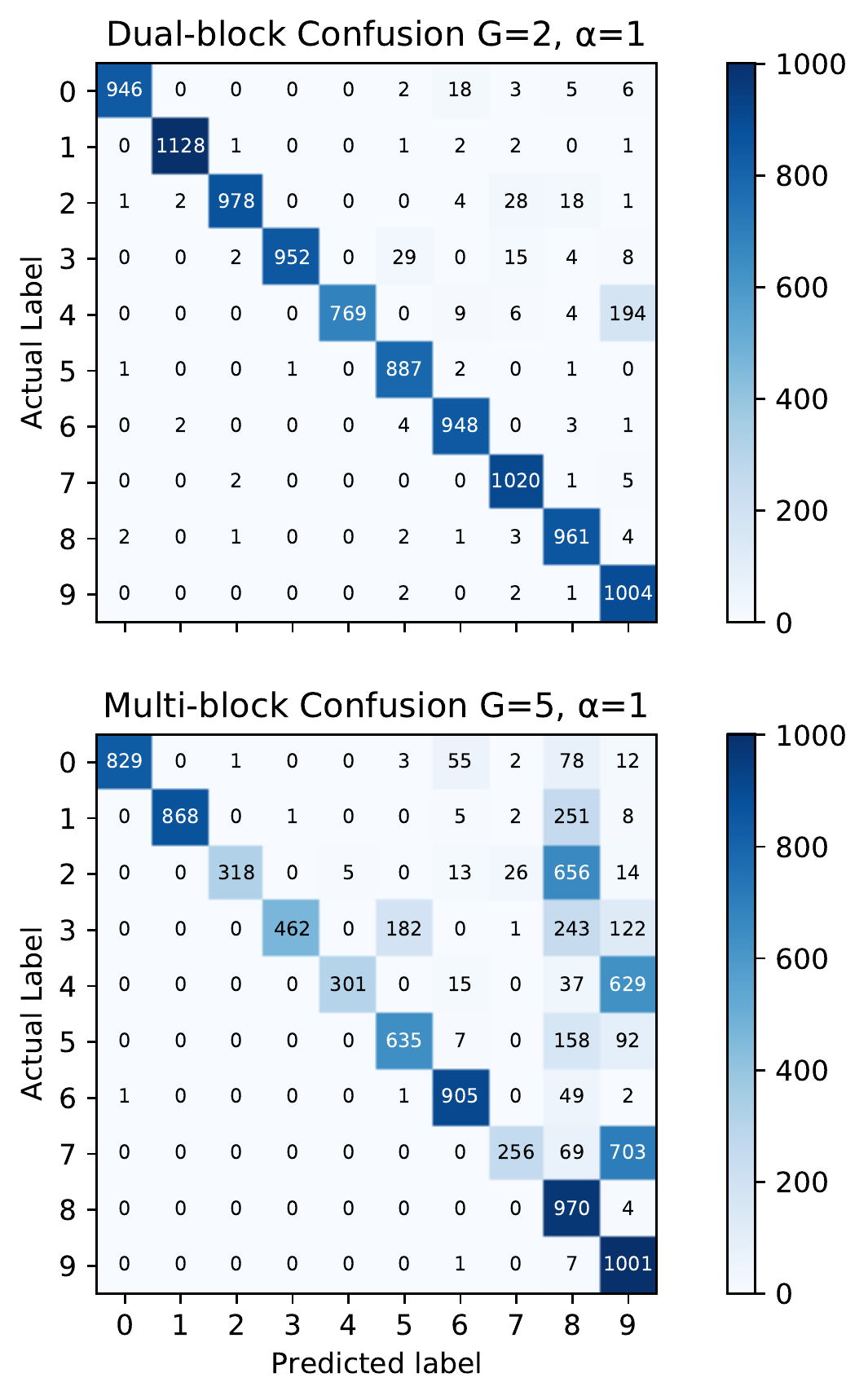}
\caption{Classification predictions of \texttt{FedAvg} regarding the block-structure. Two-block structure, where the last block that performs training holds the classes $\lbrace 5,6,7,8,9 \rbrace$. Five-block structure, where the last block that performs training holds the classes $\lbrace 8,9 \rbrace$. Caused by catastrophic forgetting, \texttt{FedAvg} treats each block like a separate task and abruptly overwrites the previously learned information once information (in the form of data) is received from the subsequent block. This is particularly noticeable in the multi-block structure.}
\label{fig:Figure 3}
\end{figure}

In Figure \ref{fig:Figure 3}, we show the confusion matrix for the predictions of the consensus model separated by two-block and five-block structure. We find that the two-block structure only shows a slight inclination towards the block of the final round, while cycling over a multi-block structure is prone to catastrophic forgetting \cite{kirkpatrick2017overcoming}. We explain this behavior by the fact that \texttt{FedAvg} treats each block as a separate task, forgetting the previous task. Although not explicitly depicted in Figure \ref{fig:Figure 3}, we find that the proneness to block-structures is increased by higher block imbalance (not illustrated).


\subsection{Convergence Analysis}

\begin{figure}[t]
\centering
\includegraphics[scale=0.5]{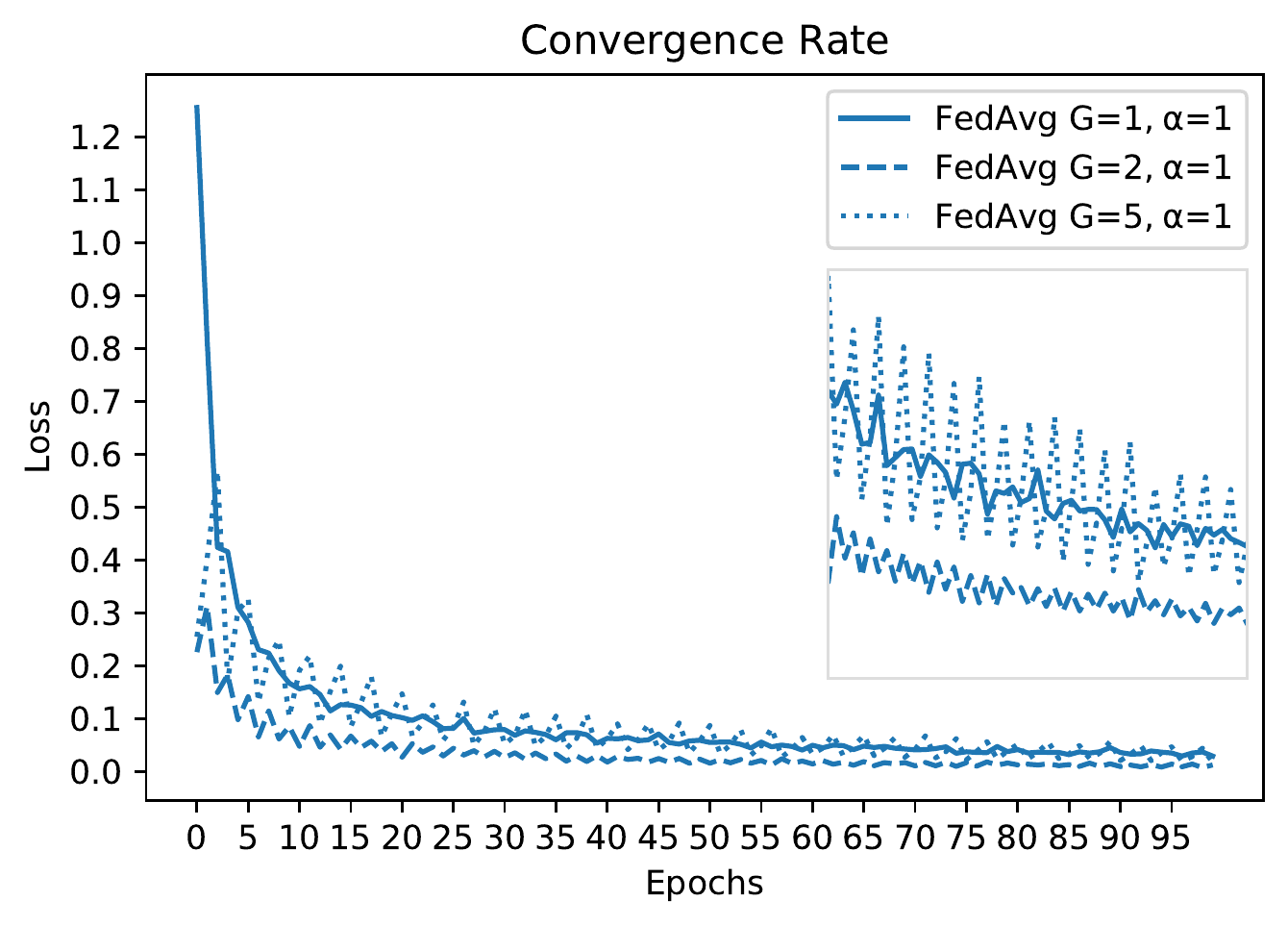}
\caption{\texttt{FedAvg} training error with block-cyclic sampling. $G=1$ represents the convergence rate for the practical upper bound for non-cyclic non-i.i.d. training. We zoom into the interval $[20,70]$ to show the oscillation of the error rate. While the two-block structure hardly oscillates, the oscillation in the training error of the multi-block structure slows down the convergence considerably.}
\label{fig:Figure 4}
\end{figure}

Following the performance analysis, we present the detailed convergence timeline graphs for balanced group distributions in Figure \ref{fig:Figure 4}. Convergence rates are plotted in increments of $5$ rounds. We observe that the training in FL is robust to cyclic sampling up to two-block structures. In addition to reduced end-of-training test accuracy, we observe more volatile training error in the case of the multi-block structure. This oscillation makes it to achieve convergence.

\begin{figure}[t]
\centering
\includegraphics[scale=0.5]{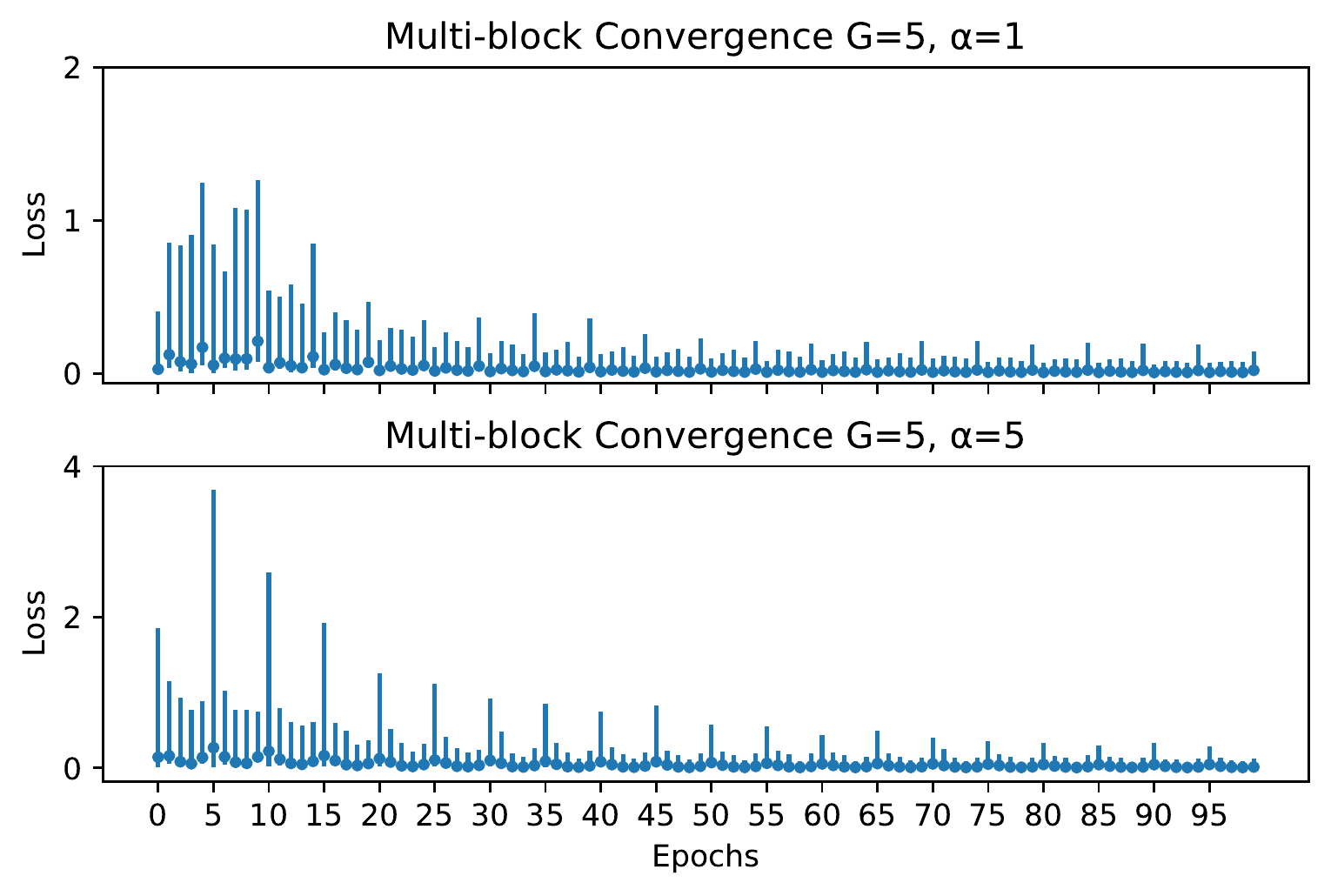}
\caption{\texttt{FedAvg} training error with multi-block cyclic sampling with respect to block imbalance $\alpha$. A value of $\alpha =1$ indicates balanced blocks. A value of $\alpha =5$ indicates unbalanced blocks. Vertical lines illustrate the highest and lowest training error achieved within the subset of selected clients per round. Note that top and bottom diagram are drawn at different scales.}
\label{fig:Figure 5}
\end{figure}

Due to space limitations, we focus our analysis of block imbalance on the five-block structure. From Figure \ref{fig:Figure 5}, we find that the convergence slowdown is even more evident when the blocks are of skewed size. The vertical lines represent the lowest and highest loss achieved by clients selected at each training round. Although the average per-round training loss is low, the vertical lines clearly indicate that the training loss varies significantly. The variation is more pronounced when learning from unbalanced blocks. From the error amplitudes occurring every $5{\text{th}}$ epoch we can clearly distinguish the training intervals of the minority block.

To quantify the slowdown introduced by cycling over heterogeneous multi-block structures, we train the consensus model until convergence. With more rounds of communication and more frequent synchronization, we could compensate for the multi-block cycling. We begin with balanced distributions $\alpha =1$. By doubling the number of communication rounds, \ie, $T=200,E=3$, we measure a consensus accuracy of $0.8878$. This represents a performance gain of about $20$\%. By synchronizing weights more frequently in addition to doubling the number of communication rounds, \ie, $T=200,E=1$, we measure a consensus accuracy of $0.9392$ which resembles approximately the value of cyclic sampling from two blocks. Considering this rather simple visual classification task, the results show that the performance decrease induced by high degrees of independence can hardly be compensated by additional optimization budget. This is even more concerning for unbalanced block distributions. For $\alpha = 5$, doubling the number of communication rounds, \ie, $T=200,E=3$, improves the test accuracy only slightly by about $18$\%. A further tripling of the synchronization resulted in no further improvement.

\subsection{Fairness Analysis}

\begin{figure}[t]
\centering
\includegraphics[scale=0.5]{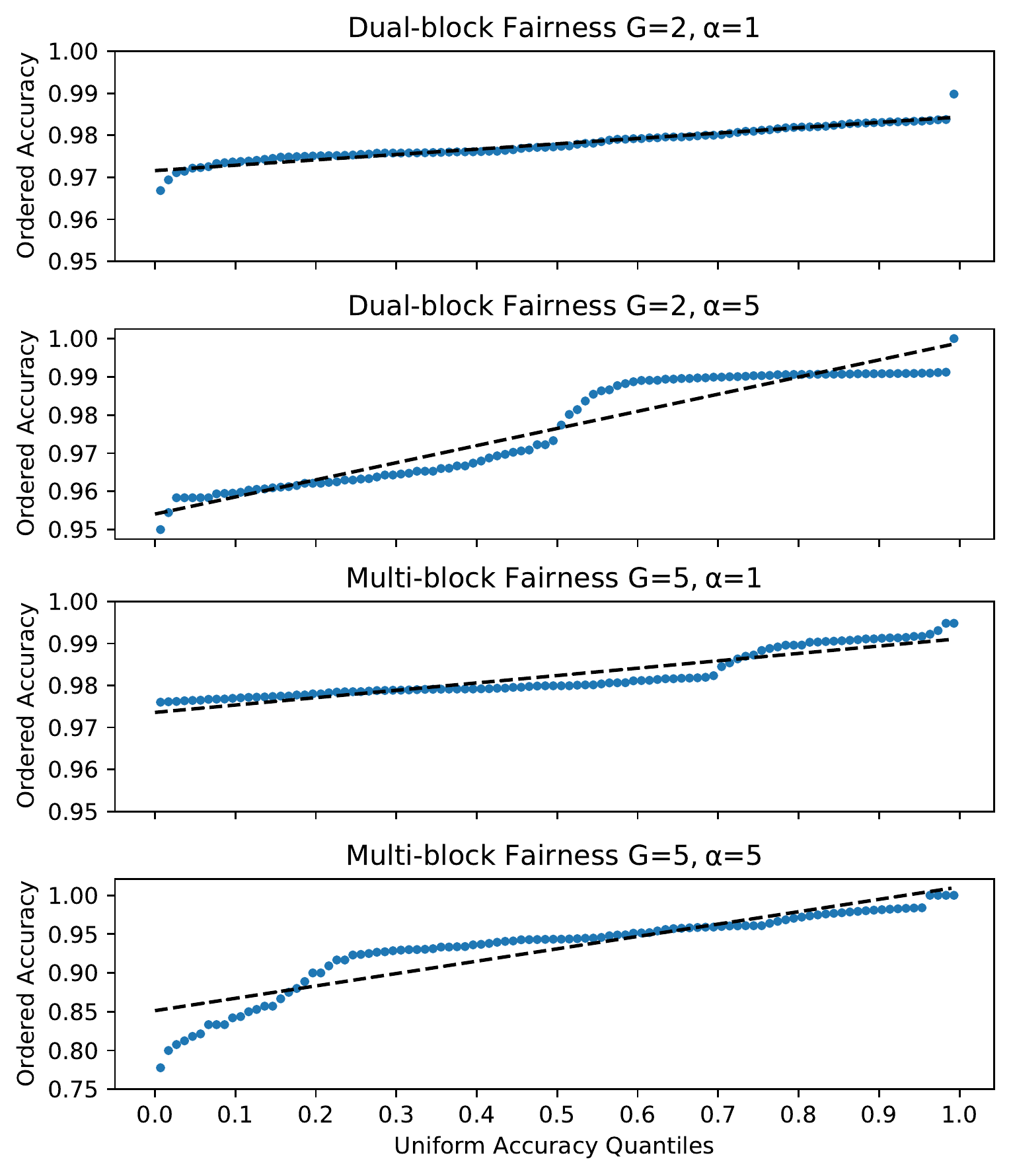}
\caption{\texttt{FedAvg} test accuracy distribution with respect to different variants of block-cyclic sampling. Figures are sorted in ascending order of complexity. The dashed line represents a theoretical uniform distribution. It resembles the fairness-optimal accuracy distribution from a client-level perspective.}
\label{fig:Figure 6}
\end{figure}

Abandoning the fixed optimization budget in terms of communication-efficiency, we unset the total number of rounds and train until convergence. By training until convergence we achieve a test accuracy of the consensus model above $0.9616$ for all experiments. To draw conclusions about the effect of block-cycling sampling on fairness, we plot the ordered local accuracies on a quantile-quantile diagram in Figure \ref{fig:Figure 6}. The dashed line represents the fairness-optimal accuracy distribution. With the exception of a few slightly underperforming clients, we find the the balanced two-block structure results in a fair distribution. With increasing block imbalance $\alpha $, we find clients from the majority block performing better with an almost constant accuracy. An in-depth analysis reveals that their performance increase does not come at the expense of clients from the minority block. We thus conclude that the performance increase can be attributed to the sampling with replacement entailing an oversampling effect. We cannot assume that multi-block structures will also lead to a fair accuracy distribution. When cycling over multi-block structures the accuracy concentrates towards the distribution tails. Having the tails heavily overpopulated while the middle is underpopulated means that a large portion of the clients will benefit less from the collaborative training. This accuracy gap is even wider for unbalanced block distributions with $\alpha >1$. In fact, the accuracy distribution of unbalanced multi-blocks deviates the most from an uniform trend.

\subsection{Sensitivity Analysis}

The learning rate is widely considered one of the most important hyperparameters, \eg, \cite{smith2017don}. For this reason, we perform a sensitivity analysis over a grid of client learning rates in logarithmic scale, \ie, $\eta \in \lbrace 0.001,0.01,0.1 \rbrace$. For the sensitivity analysis, we again reset the communication budget to $E = 3$ and $T=100$ (as in our primary analysis).

\begin{figure}[t] 
\centering
\includegraphics[scale=0.5]{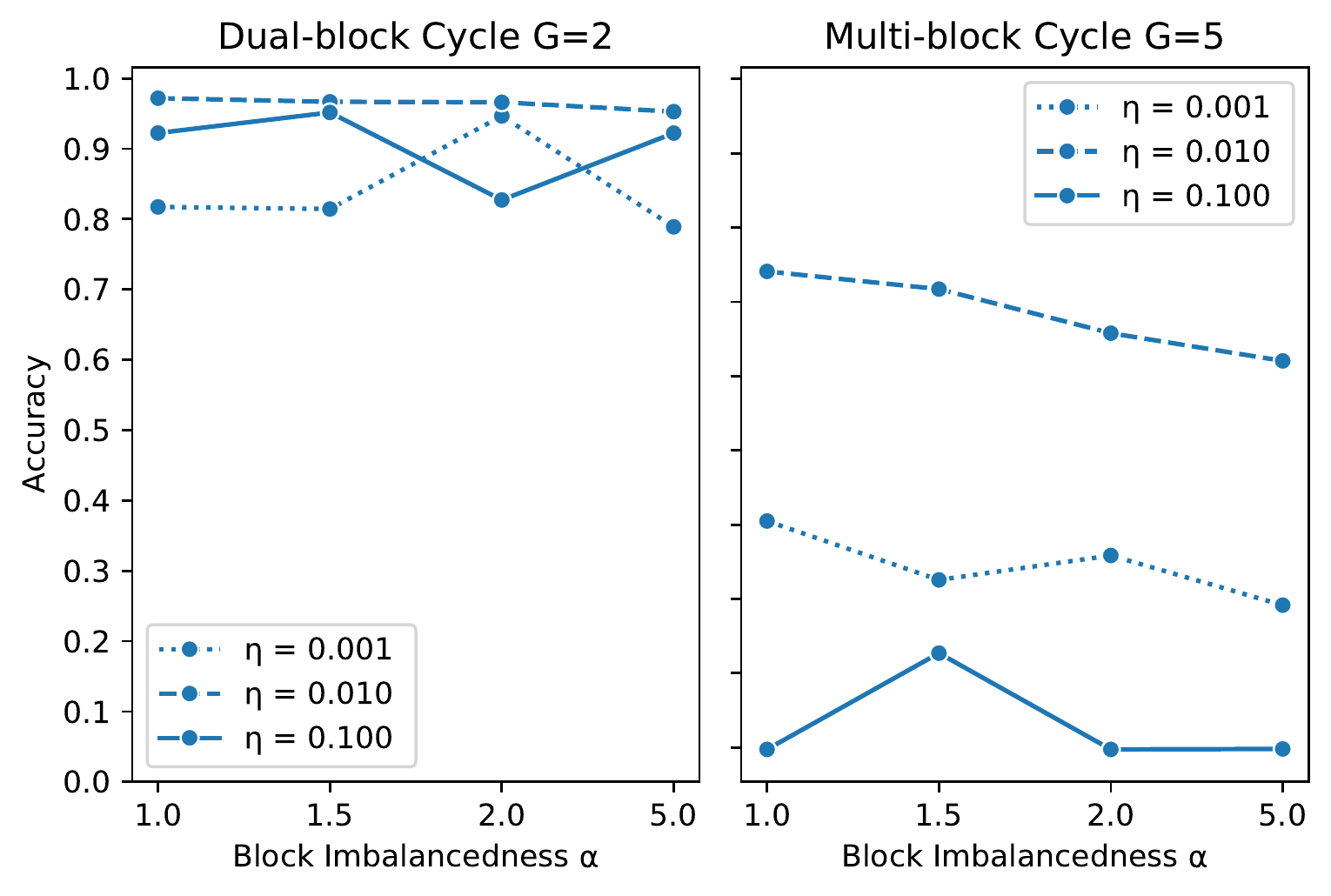} 
\caption{\texttt{FedAvg} test accuracy in terms of learning rate grouped by the ratio of dependence. The magnitude of the learning rates is encoded by the line thickness. More solid lines represent higher rates. Stronger dotted lines represent lower rates.}
\label{fig:Figure 7}
\end{figure}

From Figure \ref{fig:Figure 7}, we observe that the lines representing the test accuracy are closer together for the two-block structure compared to the five-block structure. We thus conclude that the learning rate is highly sensitive to the ratio of dependence. We also identify diminishing returns when fine-tuning the learning rates. For a small $G$, a range of learning rates (about one order of magnitude) can produce good accuracy on the holdout set. For a large $G$, careful fine-tuning of the learning rate is required to reach good accuracy. In fact, suboptimal learning rates may halve the accuracy for high ratios of dependence. We argue that learning rates that are set too high intensify catastrophic forgetting, resulting in a high misclassification rate towards the last trained block. However, once an optimal learning rate is found, it is relatively unaffected by the ratio of imbalance between blocks. 

\section{Conclusion} \label{section:5}
In response to the growing demand on the part of legislative authorities to consider user privacy, FL is increasingly employed in practice. Practical FL is exposed to statistical heterogeneity. In this study, we analyzed the effect of block-cyclic sampling embodied in FL as a result of temporal and behavioral patterns of end-users. We conduct extensive experiments on block-cyclic sampling of images for visual classification. We empirically find that the performance degrades gracefully when training with two-block structures, \eg, blocks corresponding to "day" and "night". Once block-cyclic sampling approaches a multi-block structure, significant communication overhead is required to compensate for the severe drop in classification performance. We also find that multiblock structures lead to non-uniform accuracy distributions that systematically discriminate a subset of clients. We also find that multiblock structures lead to non-uniform accuracy distributions that systematically discriminate a subset of clients.A number of avenues for future research are appealing. To further understand the impact of block-cyclic sampling in visual classification, reproducing the experiments on real-world data is required. In addition, finding methods to address the problem of data non-independence in FL while maintaining privacy and fairness is of primary concern.



{\small
\bibliographystyle{ieee_fullname}
\bibliography{references}
}

\end{document}